\documentclass{article}

\usepackage[utf8]{inputenc}

\usepackage{arxiv,amssymb,multirow,amsfonts,amsmath,graphicx,color,tikz,svg,array,makecell,csquotes,hyperref,url,booktabs,nicefrac,microtype,lipsum,natbib,doi,verbatim,listings,hyperref,caption}

\usepackage[T1]{fontenc}

\title{Retrieval-Enhanced Named Entity Recognition}

\date{}

\author{
  \href{https://orcid.org/0000-0003-4423-9278}{\includegraphics[scale=0.06]{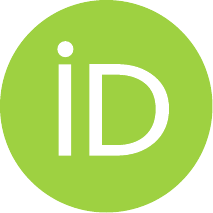}\hspace{1mm}Enzo Shiraishi} \\
  Center of Mathematics, Computing and Cognition \\
  Federal University of ABC \\
  Avenida dos Estados, 5001, Bangu, \\
  Santo André, SP, Brazil, 09.280-560 \\
  \texttt{enzo.shiraishi@aluno.ufabc.edu.br} \\
  \And
  \href{https://orcid.org/0000-0001-6021-747X}{\includegraphics[scale=0.06]{orcid.pdf}\hspace{1mm}Raphael Y.~de Camargo} \\
  Center of Mathematics, Computing and Cognition \\
  Federal University of ABC \\
  Avenida dos Estados, 5001, Bangu, \\
  Santo André, SP, Brazil, 09.280-560 \\
  \texttt{raphael.camargo@ufabc.edu.br} \\
  \And
  {Henrique L. P.~Silva} \\
  Institute of Mathematics and Statistics \\
  University of São Paulo \\
  Rua do Matão, 1010, Butantã, \\
  São Paulo, SP, Brazil, 05508-090 \\
  \texttt{henriqueluiz@usp.br} \\
  \And
  \href{https://orcid.org/0000-0001-8597-4987}{\includegraphics[scale=0.06]{orcid.pdf}\hspace{1mm}Ronaldo C.~Prati} \\
  Center of Mathematics, Computing and Cognition \\
  Federal University of ABC \\
  Avenida dos Estados, 5001, Bangu, \\
  Santo André, SP, Brazil, 09.280-560 \\
  \texttt{ronaldo.prati@ufabc.edu.br} \\
}

\renewcommand{\headeright}{}
\renewcommand{\undertitle}{}
\renewcommand{\shorttitle}{Retrieval-Enhanced Named Entity Recognition}

\hypersetup{
  pdftitle={Retrieval-Enhanced Named Entity Recognition},
  pdfsubject={cs.CL, cs.IR},
  pdfauthor={Enzo S. N.~da Silva, Raphael Y.~de Camargo, Ronaldo C.~Prati},
  pdfkeywords={Named Entity Recognition, In-Context Learning, Information Retrieval},
}
\begin{document}

\lstnewenvironment{code}[1][]
{\lstset{
    basicstyle=\ttfamily,
    firstnumber=auto,
    #1
}}
{}

\maketitle

\begin{abstract}
  When combined with In-Context Learning, a technique that enables models to adapt to new tasks by incorporating task-specific examples or demonstrations directly within the input prompt, autoregressive language models have achieved good performance in a wide range of tasks and applications. However, this combination has not been properly explored in the context of named entity recognition, where the structure of this task poses unique challenges.

  We propose RENER (Retrieval-Enhanced Named Entity Recognition), a technique for named entity recognition using autoregressive language models based on In-Context Learning and information retrieval techniques. When presented with an input text, RENER fetches similar examples from a dataset of training examples that are used to enhance a language model to recognize named entities from this input text. RENER is modular and independent of the underlying language model and information retrieval algorithms. Experimental results show that in the CrossNER collection we achieve state-of-the-art performance with the proposed technique and that information retrieval can increase the F-score by up to 11 percentage points.
\end{abstract}

\keywords{Named Entity Recognition \and In-Context Learning \and Information Retrieval}

\section{Introduction}

Autoregressive language models have achieved high performance in a wide range of tasks and applications, such as text classification and generation~(\cite{brown2020language}), using In-Context Learning (ICL). This technique allows models to adapt to new tasks by incorporating task-specific examples or demonstrations directly within the input prompt, enabling the model to learn task-specific patterns without additional fine-tuning~(\cite{min2022rethinking,dong2022survey}). Initial studies discussing the effects of the scale of language models~(\cite{wei2022emergent}) suggest that ICL is an emergent ability that scales with model size and requires a parameter count above a certain threshold. Later studies suggested alternative hypotheses, such as that the presence of emergent capabilities is caused by the model design as a whole~(\cite{lu2023emergent}) or that the existence of emergent abilities itself is related to the ICL ability of the language model~(\cite{schaeffer2024emergent}). In either case, it is clear that ICL is possible in many autoregressive language models.

ICL can offer many benefits, such as the requirement of a reduced number of supervised examples~(\cite{chowdhery2023palm}) and the possibility of implementing ICL in existing models without fine-tuning training. Being able to use unmodified underlying language models for inference allows sharing model instances with other applications, allowing an easier upgrading of the underlying language model for newer ones, an important property in this fast-evolving field. There are many studies on the use of ICL to create, optimize, and evaluate language models for many natural language processing tasks, such as text classification and generation~(\cite{min2022rethinking,brown2020language}).

Named entity recognition (NER)~(\cite{chinchor1997muc}) is a task with the objective of recognizing segments of text that refer to classes of individuals with some external meaning, called named entities~(\cite{souza2019portuguese,li2022survey,JEHANGIR2023100017}). These segments can be interpreted as mentions of these named entities in the text. For example, \enquote{John Alan Lasseter} may be a mention of a \enquote{Name} in \enquote{John Alan Lasseter is an American film director, producer, and animator}. Although there are other works with the objective of using ICL for NER~\cite{huang2020few}, their coverage is not as extensive as in other tasks~(\cite{dong2022survey,song2023comprehensive}).

There is a wide range of techniques to adapt non-autoregressive language models for NER~(\cite{souza2019portuguese,cui-etal-2021-template,zaratiana2023gliner,bogdanov2024nuner}). However, NER can sometimes be challenging for autoregressive language models because it requires understanding and generating the right segments from the input text. In other model architectures, this can also be achieved using token classification, i.e., iterating over all the tokens in the sentence and classifying the label of the current token based on the previous tokens, as in the BIO format for supervised NER annotation~(\cite{RatinovRo09}). However, since autoregressive language models are trained to predict the next token of a sequence, this technique may not be applicable using only ICL.

However, for autoregressive language models, this ability can be related to symbolic reasoning, an ability that has been shown to be significantly enhanced on some symbolic reasoning benchmarks through ICL-based reasoning techniques, such as Chain-of-Thought (CoT)~(\cite{wei2022chain}) and Scratchpad Tracing~(\cite{nye2021show}).

Besides ICL, another technique used to enhance autoregressive is retrieval-augmented generation (RAG), which uses external supporting texts selected using in formation retrieval (IR) techniques for text generation tasks such as question answering~(\cite{gao2023retrieval,ye-etal-2023-complementary}). In the context of NER, the external data could contain examples of named entities in different contexts, which the autoregressive language model could use as a reference during generation. Yet, the challenge of retrieving the most relevant examples from this external dataset still holds.

In this work, we propose RENER (Retrieval-Enhanced Named Entity Recognition), which aims to improve NER tasks using autoregressive models with the use of ICL and RAG techniques. RENER is independent of the underlying language model, does not rely on prompt engineering, and can be easily deployed for different NER domains. Experimental results show that, on the CrossNER collection~(\cite{liu2021crossner}), autoregressive language models can achieve State-of-the-Art performance using RENER and show that IR techniques can increase the F-score by up to 11 percent points when compared to equivalent systems that do not use these techniques.

\section{Related Work}

\subsection{In-Context Learning}

Because ICL techniques are mainly based on the use of templates to create prompts using examples and input data. These techniques can often be effectively combined during model adaptation. Examples include Few-Shot Learning~(\cite{brown2020language}), and Chain-of-Thought~(\cite{wei2022chain}), which use supervised annotated examples for the target task, often along with supporting data, such as explanations or demonstrations for those examples.

Min et al.~(\cite{min2022rethinking}) analyzed how autoregressive language models learn from examples using ICL, and concluded that for a wide range of models, choosing examples with input texts from the same distribution of the input text used for inference is one of the key factors to select examples for ICL that will enhance model performance. Although this conclusion encourages the study of techniques for selecting appropriate examples for ICL, the work did not evaluate every natural language processing task, instead focusing on text classification and question answering, inhibiting the possibility of directly assuming that these conclusions extend to the remaining tasks. Thus, task-specific investigations (such as in Section~\ref{section:experiments}) are necessary to extend these conclusions for NER.

\subsection{Information Retrieval}

Retrieval-augmented generation techniques~(\cite{gao2023retrieval}) have been shown to increase system performance on applications derived from text generation using information retrieval techniques to select relevant documents to be used as an additional input source for autoregressive language models. One of the hypotheses for our work is the possibility that this framework can be applied to select examples for ICL using information retrieval during inference in a way that enhances model performance.

Supporting the idea of using information retrieval to enhance ICL, Ye et al.~(\cite{ye-etal-2023-complementary}) analyzed the importance of complementary explanations to ground ICL on tasks derived from text generation, and showed that diversity in the documents plays an important role in the effectiveness of the information retrieval techniques. In the evaluated datasets, this is achieved using a method based on maximal marginal relevance~(\cite{carbonell1998use}). This conclusion emphasizes the importance of investigating the effects of information retrieval on ICL for other tasks, such as NER.

\subsection{Named Entity Recognition}

PromptNER~(\cite{ashok2023promptner}) proposed a technique to enhance the performance of ICL for NER by combining use-case-specific prompt engineering and reasoning techniques. However, the authors did not explore the effect of information retrieval techniques on ICL in the experiments that were conducted. RENER explores these possibilities and extends the work of PromptNER in many ways, and although a direct combination of both techniques is not studied in this work, it is completely possible, and this possibility is discussed in Section~\ref{section:future_works}.

GLiNER~(\cite{zaratiana2023gliner}) proposed an architecture to adapt autoencoding language models to general NER without fine-tuning. Although conceptually not requiring annotated data, the need for non-autoregressive models for NER hinders the possibility of creating systems based on ICL, which is the objective of this work. Thus, both works have different objectives.

NuNER~(\cite{bogdanov2024nuner}) proposed a method for annotating data for NER using language models and showed that RoBERTa~(\cite{liu2019roberta}) autoencoding language models fine-tuned using the data created using this process can achieve State-of-the-Art performance on the evaluated text datasets. Unlike NuNER, our proposed method does not require additional data or model fine-tuning and can work with data-scarce scenarios.

\section{RENER}
\label{section:proposal}

\begin{figure}
  \centering
  \includegraphics[width=1\linewidth]{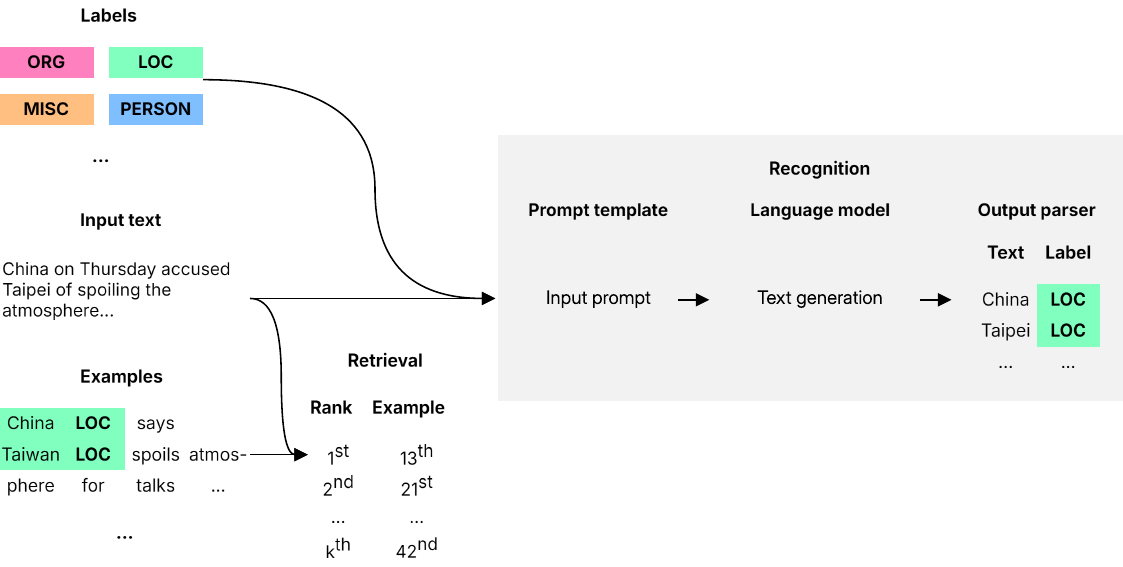}
  \caption{Flowchart representing the retrieval and recognition modules in the inference process.}
  \label{fig:overview}
\end{figure}

The objective of RENER is to combine information retrieval (IR) and In-Context Learning (ICL) techniques for NER tasks with minimal coupling, keeping the process language model agnostic. One provides RENER with an input text and obtains a list of named entities. We defined two modules: (i) a \emph{retrieval module}, responsible for selecting examples from a dataset of annotated NER examples, referred to as the training dataset; and (ii) a \emph{recognition module}, responsible for combining the selected examples and the input text into a prompt to perform NER using ICL in the autoregressive language model. Figure~\ref{fig:overview} illustrates the interaction between modules during the inference process.

\begin{figure}
  \centering
  \includegraphics{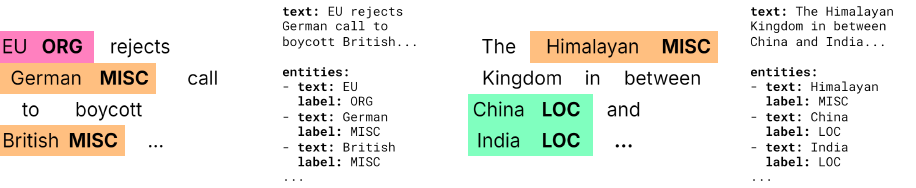}
  \caption{Example illustrations and their equivalent representations in YAML, based on samples from the dataset of the CoNLL 2003 shared task.}
  \label{fig:example}
\end{figure}

In RENER, examples consist of a text segment and a set of corresponding entities, as shown in Figure~\ref{fig:example}. Since the inference process does not rely on token-level classification, entities can be represented using a portion of the text and a string, which names the entity from a finite set of previously defined labels.

\subsection{Retrieval module}

\begin{figure}
  \centering
  \includegraphics[width=\textwidth]{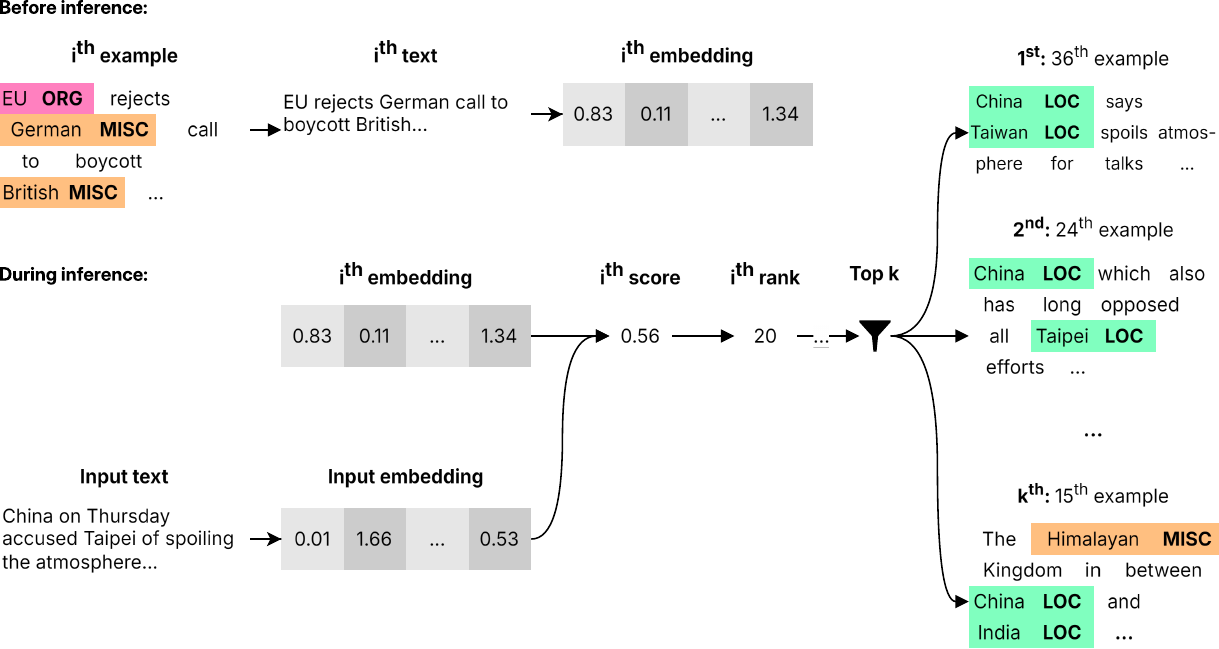}
  \caption{Flowchart representing the inference steps on the retrieval module.}
  \label{fig:retrieval}
\end{figure}

The retrieval module $\text{Rt}$ is responsible for fetching the $k$ most similar examples to an input text $t$ from the training dataset $T$, as shown in Figure~\ref{fig:retrieval}. We define the alphabet $\Sigma$, a table for the encoding standard in the application runtime, a text encoder method $\text{En}: \Sigma^* \mapsto \mathbb{R}^d$ that generates embeddings with dimentsion $d$, and a similarity score $\text{Sm}: \mathbb{R}^d \times \mathbb{R}^d \mapsto \mathbb{R}$. Since in some of the implementations described in the following sections, $\text{En}$ and $\text{Sm}$ can be inseparable, the combination of these parameters is referred to as a retrieval mechanism.

$\text{Rt}$ first maps each example from $T$ to its text and uses $\text{En}$ to encode its text into an embedding. Then, it maps each embedding to its corresponding example, creating an index $\text{In}$. This process is performed only once for each training set.

During inference, $\text{Rt}$ encodes the input text $t$ into an input embedding using $\text{En}$ and, for each example in $T$, uses $\text{Sm}$ to attribute a similarity score to $t$. $\text{Rt}$ then sorts examples based on their similarity scores and select the $k$ examples with the highest scores in order as the output. Then, this output will be used by the recognition module to name the entities in $t$. RENER can use different retrieval mechanisms, which we describe below.

\subsubsection{BM25}

BM25 is a technique used for scoring the similarity of texts based on a probabilistic framework for information retrieval~(\cite{robertson1995okapi,robertson2009probabilistic}). BM25 calculates this score using factors such as term frequency, inverse document frequency, and document length. This technique is simple to scale and interpret and is adopted in many important search engine frameworks and applications, such as Apache Lucene~(\cite{perez2009integrating}). There are multiple variations of BM25~(\cite{trotman2014improvements,10.1007/978-3-030-45442-5_4}), and in this work, for simplicity, we use the Okapi BM25 formulation~(\cite{trotman2014improvements}).

To define a BM25 retrieval mechanism, two key parameters must be specified: \( k_1 \) and \( b \), where \( k_1, b \in \mathbb{R} \). Formally, \( k_1 \) represents the term frequency saturation parameter, which controls the influence of term frequency on the final score by acting as a scaling factor for the term matches. A higher \( k_1 \) value increases the impact of term frequency, while a lower value limits it. The parameter \( b \) serves as a length normalization factor, determining how much document length affects the scoring. Specifically, \( b \in [0, 1] \), with \( b = 0 \) indicating no length normalization and \( b = 1 \) applying full normalization. This balances the relevance of documents of varying lengths by penalizing overly long documents. Both parameters, \( k_1 \) and \( b \), significantly impact the final retrieval scores, and their optimal values depend on the specific application. In practice, the default values of \( k_1 = 1.2 \) and \( b = 0.75 \) are commonly used, offering robust performance in many information retrieval tasks in the absence of hyperparameter tuning~(\cite{perez2009integrating,manning2009introduction}).

One can also apply the stemming and lemmatization normalization techniques, which have been shown to increase BM25 performance on some benchmarks~(\cite{trotman2014improvements,10.1007/978-3-030-45442-5_4}) since they reduce the number of terms present in the vocabulary and increase the matches. Stemming refers to the process of reducing words to their base or root form by removing suffixes (e.g., \enquote{running} becomes \enquote{run}), while lemmatization reduces words to their canonical dictionary form based on context (e.g., \enquote{better} becomes \enquote{good}). Since inflection removal modifies the target texts, we should apply them to a separate copy of the training dataset used only by the retrieval module.

\subsubsection{Semantic search}

In semantic search, text embedding models serve as encoders. The similarity between two texts is measured by the inverse of a distance metric between embeddings. Since these embeddings are dense vectors, each value in the vector relates to all terms in the original text, preventing the direct interpretation of scores in terms of term matches. Instead, embedding models are typically trained to ensure that the distances between the vectors correspond to the semantic similarity between the texts. This is achieved through specialized training techniques, as seen in models like SBERT~(\cite{reimers2019sentence}).

Beyond the definition of the encoder $\texttt{En}$ and the similarity score $\texttt{Sm}$, it is possible to add additional layers to semantic search, using techniques such as maximal marginal relevance~(\cite{carbonell1998use}) and reciprocal rank fusion~(\cite{cormack2009reciprocal}).

\subsubsection{Maximal Marginal Relevance}

Maximal marginal relevance~(\cite{carbonell1998use}) is a technique used to reduce the redundancy between selected examples by balancing variety and similarity between examples with the highest scores during inference. Introducing variety in retrieval-augmented generation techniques seems to enhance the generalization capabilities of text embedding models by helping to avoid biases acquired during the training steps~(\cite{ye-etal-2023-complementary}).

To define a maximal marginal relevance layer, it is necessary to specify $\lambda \in [0,1]$, a regularization parameter that controls the balance between diversity and similarity. Defining $\lambda = 0$ is the same as not using maximal marginal relevance, and higher values of $\lambda$ will increase the diversity in the outputs of the retrieval mechanism.

\subsubsection{Reciprocal Rank Fusion}
\label{section:rrf}

Reciprocal Rank Fusion (RRF)~(\cite{cormack2009reciprocal}) is a technique that combines the ranked outputs of multiple retrieval mechanisms into a single result. An example ranked highly on the final list implies that it was highly ranked in multiple individual retrieval mechanisms. RRF relies solely on the ranking positions of the combined mechanisms, which makes it independent of their specific implementations. It is used in many applications to implement what is known as a hybrid search by combining term match-based techniques (such as BM25) and semantic search techniques. Since both approaches can have their shortcomings in some applications, a hybrid search can increase the efficiency of the retrieval module in cases that demand a balance of term-matching and semantic similarity to achieve the best order for the example texts in relation to the input texts.

To define a reciprocal rank fusion layer, it is necessary to specify $C \in \mathbb{R}$, a regularization parameter for the order of elements after scoring. This parameter also has a significant effect on the final scores, and its optimal value is application-specific. However, in the absence of a hyperparameter optimization for these values, $C = 60$ is found to be the value with the best average result in the benchmarks evaluated in the original work~(\cite{cormack2009reciprocal}).

\subsection{Recognition module}

\begin{figure}
  \centering
  \includegraphics[width=\textwidth]{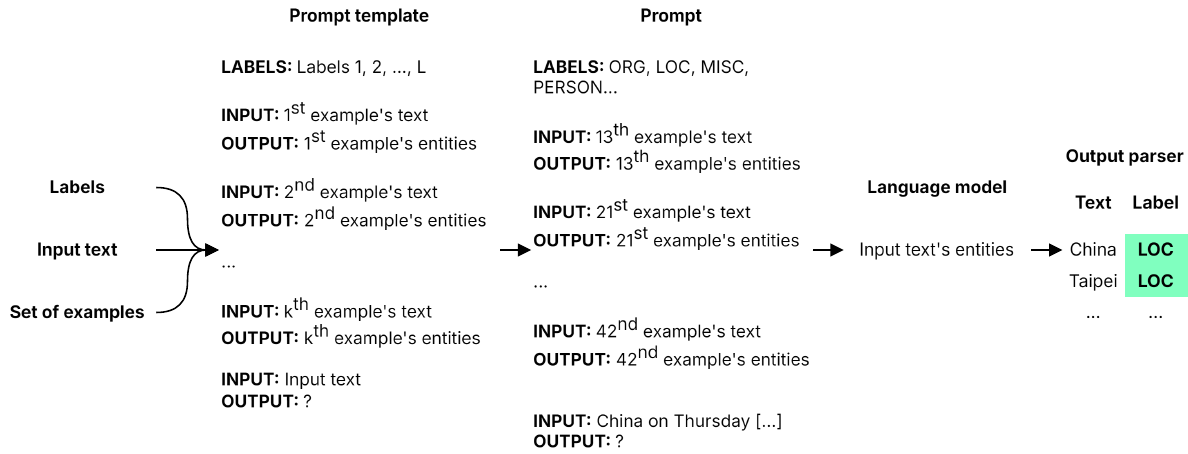}
  \caption{Flowchart representing the inference steps on the recognition module.}
  \label{fig:recognition}
\end{figure}

The recognition module combines the input text and a set of examples from the retrieval module and uses an autoregressive language model and output parser to recognize named entities. The input sources are first transformed into a prompt using a template, which is used as the sole input of the language model. The model generates the output text containing the predicted entities, which is converted into a set of entities by an output parser.

The parser uses regular expressions to transform the text into objects, which guarantees that all parsed entities have valid names and that their segments are part of the input text. It also allows the same procedures to be used to evaluate RENER and other NER techniques.

\subsection{Evaluation}

NER models can be evaluated using the same evaluation metrics used to measure the efficiency of many text classification models, such as accuracy, precision, recall, and F-score with some adaptations~(\cite{JEHANGIR2023100017}). However, instead of comparing predicted and expected classes between samples, the intersection of the predicted and expected entity sets is analyzed, taking into account the value and label of each entity. True positives are obtained from entities that belong to both sets. False positives are obtained from entities that belong to the predicted set, but not to the expected set, and false negatives are obtained from the entities that belong to the expected set, but not to the predicted set.

\section{Experiments}

In this section, we present the RENER experimental evaluation procedure.

\subsection{Data}

\begin{table}
  \centering
  \begin{tabular}{|l|l|>{\raggedright\arraybackslash}p{0.3\linewidth}|r|r|r|}
    \hline
    \textbf{Domain} & \textbf{Source} &  \textbf{Labels}&\textbf{\makecell[l]{ \\ Training \\ samples \\ ~}}& \textbf{\makecell[l]{ \\ Validation \\ samples \\  ~}} & \textbf{\makecell[l]{ \\ Test \\ samples \\  ~}} \\

    \hline
    Reuters &CoNLL 2003 & \texttt{person, organization, location, miscellaneous}&14,987  &3,466 &3,684 \\
    \hline
    Politics &Wikipedia & \texttt{politician, person, organization, political party, event, election, country, location, miscellaneous}&200  &541 &651 \\
    \hline
    Natural Science &Wikipedia & \texttt{scientist, person, university, organization, country, location, discipline,
      enzyme, protein, chemical compound, chemical element, event,
    astronomical object, academic journal, award, theory, miscellaneous}&200  &450 &543 \\
    \hline
    Music &Wikipedia & \texttt{music genre, song, band, album, musical artist, musical instrument,
    award, event, country, location, organization, person, miscellaneous}&100  &380 &456 \\
    \hline
    Literature &Wikipedia & \texttt{book, writer, award, poem, event, magazine, person, location,
    organization, country, miscellaneous}&100  &400 &416 \\
    \hline
    Artificial Intelligence &Wikipedia & \texttt{field, task, product, algorithm, researcher, metrics, university,
    country, person, organization, location, miscellaneous}&100  &350 &431 \\
    \hline
  \end{tabular}

  \caption{Details for the datasets composing the CrossNER collection}
  \label{table:crossner}
\end{table}

We used mainly the CrossNER dataset, a cross-domain collection of English NER datasets focused on evaluating the ability of models in situations of training data scarcity. CrossNER comes from five domains, with between 100 and 200 training samples obtained from Wikipedia from a wide range of subjects. Although training data are scarce, there are at least three times more validation and test samples, allowing a proper model evaluation. Furthermore, to mitigate subjectivity in the entity annotation process, all entities are based on the consensus of multiple human specialists during the annotation process~(\cite{liu2021crossner}). Table~\ref{table:crossner} describes the characteristics of each domain in detail.

We also used the NER dataset from the CoNLL-2003 shared task obtained from Reuters news stories collected from many topics between August 1996 and August 1997~(\cite{tjong-kim-sang-de-meulder-2003-introduction}). It has a very large number of training samples, allowing evaluation of performance in situations where large amounts of training data are available, but the domain of the application is broader, requiring more generalization from the language model. It also eases comparison with the results of other general-purpose NER techniques since this dataset has been widely used as a benchmark.

\subsection{Parameters}
\label{section:parameters}

We evaluated RENER using the three implementations of the retrieval mechanism (BM25, semantic, and hybrid search) on each dataset in the CrossNER collection~(\cite{liu2021crossner}). For each implementation, we performed experiments using all $k \in [1,25]$ values to evaluate the impact of the number of examples in the prompt.

For standardization, when using BM25, $k_1$ and $b$ are always defined as $k_1 = 1.2$ and $ b = 0.75$. Additionally, lemmatization is used to process a separate copy of every training dataset using the \enquote{lemmatizer} pipeline from the \enquote{en\_core\_web\_lg} package on the spaCy framework~(\cite{honnibal2020spacy}), which is English-specific and rule-based. Thus, this step is only present during the retrieval mechanism inference and will not interfere with the inference of other modules.

When using semantic search, \enquote{bge-large-en-v1.5}, an English general purpose fine-tuned RoBERTa model~(\cite{chen2024bge,liu2019roberta}), is used as the encoder, and the cosine similarity function (or the inverse of the cosine distance) is used as the distance metric. These choices are based on the reported performance of this text embedding model on the Massive Text Embed Benchmark~(\cite{muennighoff2022mteb}) for text retrieval tasks in English. This model can also take query instructions. To guarantee the equality between both studies, we use the same query instruction \enquote{Represent this sentence for searching relevant passages:}~(\cite{chen2024bge}). We used maximal marginal relevance with $\lambda = 0$ as a baseline and $\lambda = 0.5$ for the scenario where diversity and similarity are considered equally important.

When using hybrid search, we combine the baseline semantic search and BM25 using reciprocal rank fusion, and $C$ is always defined as $C = 60$. Finally, models that do not use any form of dynamic example selection use the same randomly selected $k$ examples on every inference.

Two autoregressive language models are evaluated during experiments: Gemini (\enquote{gemini-1.5-pro})~(\cite{team2023gemini}), and GPT-4o (\enquote{gpt-4o-2024-05-13})~(\cite{achiam2023gpt}). This choice is based on the reported performance of these models on the Massive Multitask Language Understanding~(\cite{hendrycks2020measuring}) and Grade School Math 8K~(\cite{cobbe2021training}) benchmarks, which evaluate performance on tasks derived from text generation that demand from the model high ICL and reasoning capabilities.

The same prompt template is used in every inference (shown next), and it is an adaptation of PromptNER~(\cite{ashok2023promptner}) achieved by abstracting declarations of use-case specific instructions and non-entity spans on example definitions (used originally as a reasoning technique). This choice strives to allow replicating experiments in other NER datasets without requiring additional annotations, as well as to allow evaluating the contribution of information retrieval and reasoning techniques separately, which Section~\ref{section:future_works} describes.

The source code, templates and parameters used for reproduction purposes can also be found in the official repository of RENER\footnote{\url{https://github.com/eshiraishi/rener}}.

Finally, to compare the performance of these models with the results of models using other architectures, the results of the model with State-of-the-Art performance on the CrossNER collection~(\cite{liu2021crossner}), \enquote{NuNER-zero-span}~(\cite{bogdanov2024nuner}), are also measured.

\begin{code}
  Defn: An entity is a label 1, label 2, ... , or label L.

  Example 1: example text 1
  Answer:
  1. example 1, entity segment 1 (example 1, entity label 1)
  2. example 1, entity segment 2 (example 1, entity label 2)
  ...
  S. example 1, entity segment S (example 1, entity label S)


  Example 2: example text 2
  Answer:
  1. example 2, entity segment 1 (example 2, entity label 1)
  2. example 2, entity segment 2 (example 2, entity label 2)
  ...
  T. example 2, entity segment T (example 2, entity label T)
  ...
  Example E: example text E
  Answer:
  1. example E, entity segment 1 (example E, entity label 1)
  2. example E, entity segment 2 (example E, entity label 2)
  ...
  U. example E, entity segment U (example E, entity label U)

  Example E+1: input text
  Answer:
\end{code}

\subsection{Results}

In the experiments, we initially determine the best $k$ values for each IR mechanism using the validation dataset, followed by the use of the selected configuration using the test dataset.

\subsubsection{Hyperparameter optimization}
\label{section:experiments}

\begin{figure}
  \centering
  \includegraphics[width=0.8\linewidth]{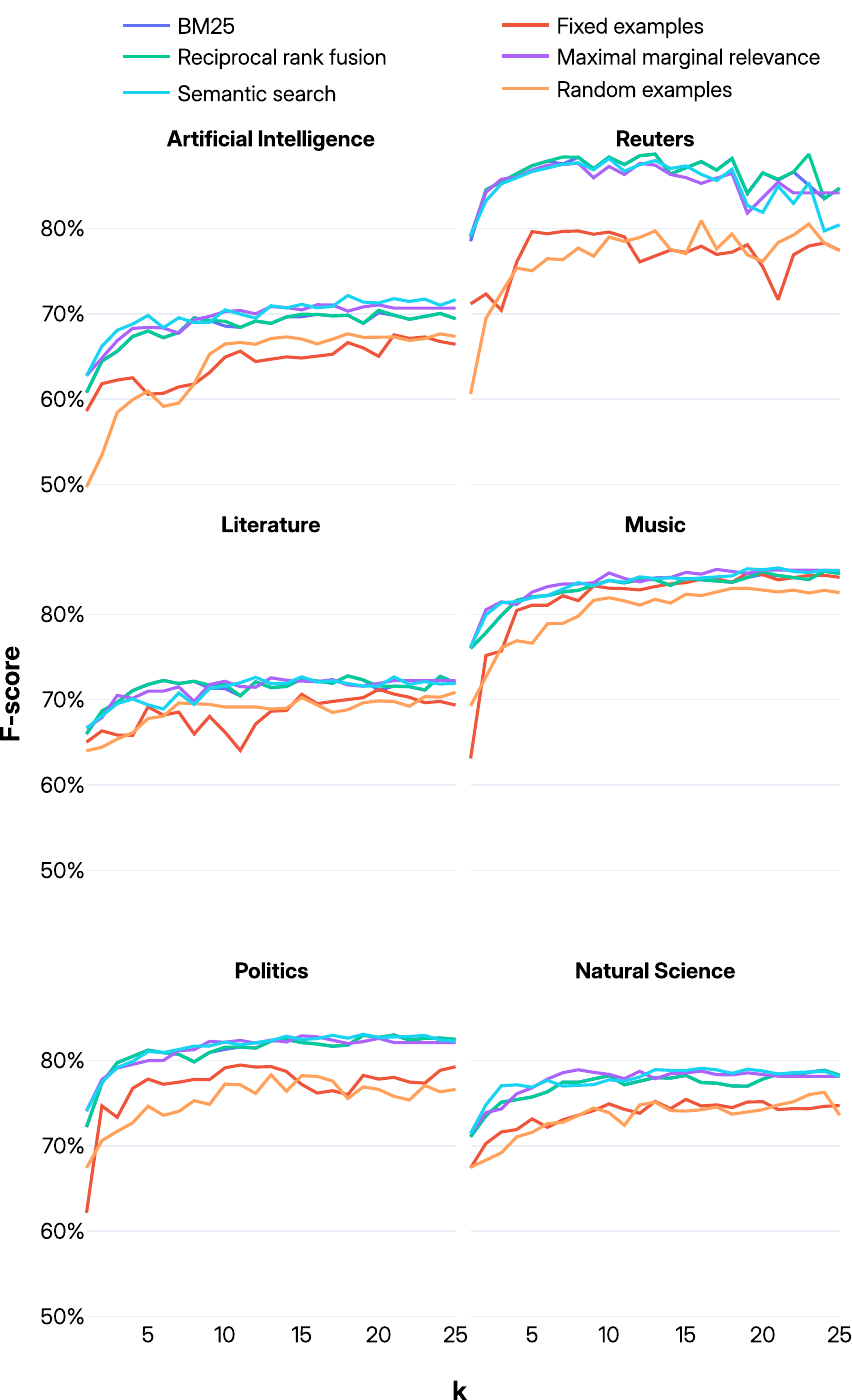}
  \caption{Chart relating the F-score and $k$ on the validation datasets of CrossNER for all retrieval mechanisms using Gemini.}
  \label{fig:results:validation}
\end{figure}

We evaluated RENER using Gemini and the validation datasets for all retrieval mechanisms to determine their relative performance as we increase the value of $k$. Figure~\ref{fig:results:validation} shows
that the usage of information retrieval techniques generally increases the F-score of the language model compared to selecting fixed and random examples. Table~\ref{table:results:validation} details the configurations with the best F-score for every dataset in the best-case scenario, showing that information retrieval techniques increase the F-score by up to 11 percent and have improvements in all scenarios.

\begin{table}
  \centering
  \begin{tabular}{|l|r|l|r|r|}
    \hline
    \textbf{Domain} & \textbf{$k$} & \textbf{Retrieval mechanism}& \textbf{F-score} (using retrieval)& \textbf{F-score} (without retrieval)\\
    \hline
    Artificial Intelligence& 18 & Semantic search& 72.14\%  &66.62\%\\
    Literature & 18 & BM25 & 72.78\%  &70.00\%\\
    Music & 21 & Semantic search& 85.45\%  &84.04\%\\
    Natural Science& 16 & Semantic search& 79.09\%  &74.67\%\\
    Politics & 19 & Semantic search& 83.07\%  &78.00\%\\
    Reuters & 13 & BM25 & 88.71\%  &77.00\%\\
    \hline
  \end{tabular}
  \caption{Details from the models with best results for every validation datasets of CrossNER using Gemini}
  \label{table:results:validation}
\end{table}

Although baseline semantic search and BM25 are present in all the best results in the validation datasets (Table~\ref{table:results:validation}), hybrid search and maximal marginal relevance show comparable results in most domains (Figure~\ref{fig:results:validation}). Since we did not optimize their hyperparameters, their results could still be improved. Nevertheless, although no single IR technique was distinctly superior, all of them clearly improved the F-score in comparison to random or fixed choice of examples. This shows the importance of using a proper IR technique to select examples for ICL in NER applications.

\subsubsection{Evaluation}

\begin{figure}
  \centering
  \includegraphics[width=\linewidth]{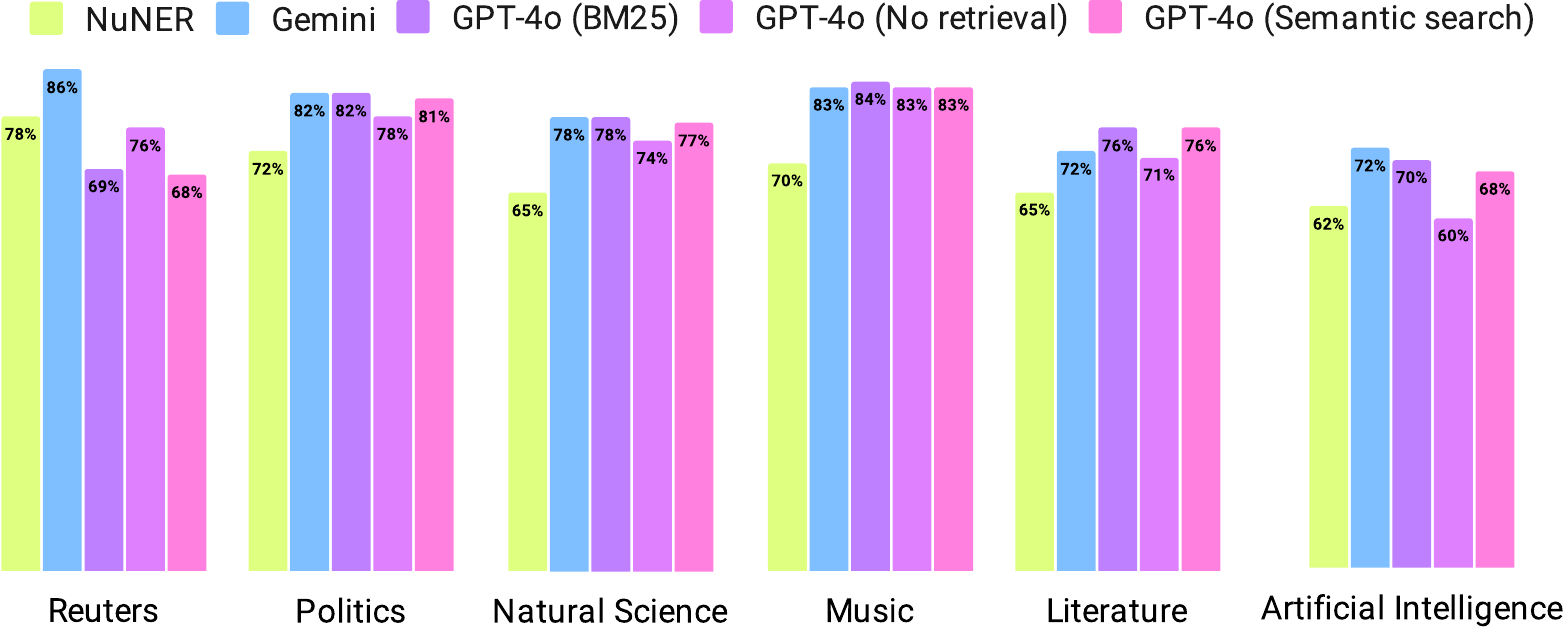}
  \caption{Chart comparing, for the test datasets of CrossNER, the F-score of NuNER and RENER (using Gemini with the parameters from Table~\ref{table:results:validation}) and GPT-4o (using either no retrieval, semantic search or BM25, and $k$ = 20)}
  \label{fig:results:test}
\end{figure}

\begin{table}
  \centering
  \begin{tabular}{|l|r|l|r|r|}
    \hline
    \textbf{Domain} & \textbf{$k$} & \textbf{Retrieval mechanism}& \textbf{F-score} (validation dataset)& \textbf{F-score} (test dataset)\\
    \hline
    Artificial Intelligence& 18 & Semantic search& 72.14\%  &72.09\%\\
    Literature & 18 & BM25 & 72.78\%  &72.11\%\\
    Music & 21 & Semantic search& 85.45\%  &83.21\%\\
    Natural Science& 16 & Semantic search& 79.09\%  &78.43\%\\
    Politics & 19 & Semantic search& 83.07\%&82.02\%\\
    Reuters & 13 & BM25 & 88.71\%  &86.10\%\\
    \hline
  \end{tabular}
  \caption{Comparison of the F-score of the models with best results for the validation and test datasets of CrossNER using Gemini}
  \label{table:results:test}
\end{table}

We executed the Gemini RENER implementation on the CrossNER test datasets using the configurations that obtained the best results in the validation sets. We obtained F-scores that were very similar to those obtained with the validation set (Table \ref{table:results:test}), with the largest difference being in the Reuter dataset, with 2.6\%.

To compare these values with the performance of other language models, we also measure the results when using GPT-4o. To reduce the usage of computational resources, we only evaluated the BM25 and semantic search mechanisms, which had the best results in Gemini with the validation dataset. We also adopt a single $k = 20$ number of examples, which was in line with the best results in Gemini. Finally, we compare the results with \enquote{NuNER-zero-span}~(\cite{bogdanov2024nuner}), which has state-of-the-art performance on the CrossNER collection.

Figure~\ref{fig:results:test} shows that RENER models perform better in all domains in the CrossNER dataset compared to NuNER, showing that RENER can effectively work with scarce data. Moreover, GPT-4o performs similarly to Gemini in the CrossNER domains, even though there was no parameter fine-tuning for the former. In the Reuters domain, only the Gemini model was clearly superior to NuNER, which could be because for this domain, a small $k$ value of 13 was optimal, while performance degraded with more examples (Figure~\ref{fig:results:validation}). This dataset is also the only case where GPT-4o performed better with no retrieval mechanisms, though it did not perform better than Gemini using a retrieval mechanism.

\section{Conclusion}

\label{section:future_works}

In this work, we proposed a novel technique to enhance the performance of autoregressive language models for NER using ICL information retrieval techniques. The experiments showed that in almost all situations, information retrieval increased the performance of autoregressive language models on NER when using ICL. It should be noted that even though the language models were used with no modification or fine-tuning, we achieved State-of-the-Art performance on the CrossNER collection, showing that RENER can be an effective alternative for NER in situations of training data scarcity. It is important to mention that GPT-4o obtained results comparable to Gemini even without any parameter optimization, which further supports the use of these techniques in data-scarce scenarios.

This work can be further extended by performing parameter optimization for the information retrieval mechanisms using hyperparameter optimization techniques. This investigation could also allow discussion of how independent these parameters are in relation to the underlying language model. Another direction is the evaluation of other language models and other datasets to determine if the obtained results hold in more general scenarios. Furthermore, this can help in the discussion if these results are affected by model contamination caused by the presence of the validation and test datasets in the training data of the language models, a concern first proposed in PromptNER~(\cite{ashok2023promptner}).

To extend the proposed architecture, it is possible to evaluate the effects of integrating proper reasoning techniques as part of the recognition module, such as CoT~(\cite{wei2022chain}) through the adaptation proposed in PromptNER~(\cite{ashok2023promptner}). The need for additional annotations over existing datasets can potentially be mitigated by evaluating the feasibility of improving the data annotation process using synthetic data generation processes, such as the techniques proposed in \enquote{Auto-CoT}~(\cite{zhang2023automatic}) and NuNER~(\cite{bogdanov2024nuner}).

\bibliographystyle{unsrtnat}
\bibliography{main}

\end{document}